\documentclass[conference]{IEEEtran}
\ifCLASSINFOpdf
\else
\fi

\usepackage{graphicx}
\usepackage{algorithm}
\usepackage[noend]{algorithmic}
\usepackage{qtree}
\usepackage{color}
\usepackage[utf8]{inputenc} % set input encoding (not needed with XeLaTeX)
\usepackage[table]{xcolor}

\begin{document}
%
% paper title
% can use linebreaks \\ within to get better formatting as desired
\title{Automatic Wrapper Adaptation by\\ Tree Edit Distance Matching}

% author names and affiliations
% use a multiple column layout for up to three different
% affiliations
\author{\IEEEauthorblockN{Emilio Ferrara}
\IEEEauthorblockA{Department of Mathematics\\University of Messina, Italy\\
Email: emilio.ferrara@unime.it}
\and
\IEEEauthorblockN{Robert Baumgartner}
\IEEEauthorblockA{Lixto Software GmbH\\Vienna, Austria\\
Email: robert.baumgartner@lixto.com}
}

% conference papers do not typically use \thanks and this command
% is locked out in conference mode. If really needed, such as for
% the acknowledgment of grants, issue a \IEEEoverridecommandlockouts
% after \documentclass

% for over three affiliations, or if they all won't fit within the width
% of the page, use this alternative format:
% 
%\author{\IEEEauthorblockN{Michael Shell\IEEEauthorrefmark{1},
%Homer Simpson\IEEEauthorrefmark{2},
%James Kirk\IEEEauthorrefmark{3}, 
%Montgomery Scott\IEEEauthorrefmark{3} and
%Eldon Tyrell\IEEEauthorrefmark{4}}
%\IEEEauthorblockA{\IEEEauthorrefmark{1}School of Electrical and Computer Engineering\\
%Georgia Institute of Technology,
%Atlanta, Georgia 30332--0250\\ Email: see http://www.michaelshell.org/contact.html}
%\IEEEauthorblockA{\IEEEauthorrefmark{2}Twentieth Century Fox, Springfield, USA\\
%Email: homer@thesimpsons.com}
%\IEEEauthorblockA{\IEEEauthorrefmark{3}Starfleet Academy, San Francisco, California 96678-2391\\
%Telephone: (800) 555--1212, Fax: (888) 555--1212}
%\IEEEauthorblockA{\IEEEauthorrefmark{4}Tyrell Inc., 123 Replicant Street, Los Angeles, California 90210--4321}}

% use for special paper notices
%\IEEEspecialpapernotice{(Invited Paper)}

% make the title area
\maketitle

\begin{abstract}
%\boldmath
Information distributed through the Web keeps growing faster day by day, and for this reason, several techniques for
extracting Web data have been suggested during last years. Often, extraction tasks are performed through so called
wrappers, procedures extracting information from Web pages, e.g. implementing logic-based techniques. Many fields of
application today require a strong degree of robustness of wrappers, in order not to compromise assets of information
or reliability of data extracted. Unfortunately, wrappers may fail in the task of extracting data from a Web page, if
its structure changes, sometimes even slightly, thus requiring the exploiting of new techniques to be automatically
held so as to adapt the wrapper to the new structure of the page, in case of failure. In this work we present a novel
approach of \emph{automatic wrapper adaptation} based on the measurement of similarity of trees through improved tree
edit distance matching techniques.
\end{abstract}
% IEEEtran.cls defaults to using nonbold math in the Abstract.
% This preserves the distinction between vectors and scalars. However,
% if the conference you are submitting to favors bold math in the abstract,
% then you can use LaTeX's standard command \boldmath at the very start
% of the abstract to achieve this. Many IEEE journals/conferences frown on
% math in the abstract anyway.

% no keywords

% For peer review papers, you can put extra information on the cover
% page as needed:
% \ifCLASSOPTIONpeerreview
% \begin{center} \bfseries EDICS Category: 3-BBND \end{center}
% \fi
%
% For peerreview papers, this IEEEtran command inserts a page break and
% creates the second title. It will be ignored for other modes.
\IEEEpeerreviewmaketitle

\section{Introduction}
% no \IEEEPARstart

Web data extraction, during last years, captured attention both of academic research and enterprise
world because of the huge, and still growing, amount of information distributed through the Web. Online documents are
published in several formats but previous work primarily focused on the extraction of information from HTML Web pages.

Most of the wrapper generation tools developed during last years provide to full support for users in building data extraction programs (a.k.a.\ wrappers) automatically and in a visual
way. They can reproduce the navigation flow simulating the human behavior, providing support for technologies adopted to develop Web pages, and so on. Unfortunately, a problem still holds: wrappers, because of their intrinsic nature and the complexity of extraction tasks they perform, usually are strictly connected to the structure of Web pages (i.e. DOM tree) they handle. Sometimes, also slight changes to that structure can cause the failure of extraction tasks. A couple of wrapper generation systems try to natively avoid problems caused by minor changes, usually building more elastic wrappers (e.g. working with relative, instead of absolute, XPath queries to identify elements).

Regardless of the degree of flexibility of the wrapper generator, wrapper maintenance is still a required step of a wrapper life-cycle. Once the wrapper has been correctly developed, it could work for a long time without any malfunction. The main problem in the wrapper maintenance is that no one can predict when or what kind of changes could occur in Web pages.

Fortunately, local and minor changes in Web pages are much more frequent case than deep modifications (e.g.
layout rebuilding, interfaces re-engineering, etc.). However, it could also be possible, after a minor modification on
a page, that the wrapper keeps working but data extracted are incorrect; this is usually even worse, because it causes
a lack of consistency of the whole data extracted. For this reason, state-of-the-art tools started to perform
validation and cleansing on data extracted; they also provide caching services to keep copy of the last working version
of Web pages involved in extraction tasks, so as to detect changes; finally, they notify to maintainers any change,
letting possible to repair or rewrite the wrapper itself. Depending on the complexity of the wrapper, it could be more
convenient to rewrite it from scratch instead of trying to find causes of errors and fix them.

Ideally, a robust and reliable wrapper should include directives to auto-repair itself in case of malfunction or
failure in performing its task. Our solution of automatic wrapper adaptation relies on exploiting the possibility of
comparing some structural information acquired from the old version of the Web page, with the new one, thus making it
possible to re-induct automatically the wrapper, with a custom degree of accuracy.

The rest of the paper is organized as follows: in Section 2 we consider the related work on theoretical background and Web data extraction, in particular regarding algorithms, techniques and problems of wrapper maintenance and adaptation. Sections 3 covers the automatic wrapper adaptation idea we developed, detailing some interesting aspects of algorithms and providing some examples. Experimentation and results are discussed in Section 4. Section 5, finally, presents some conclusive considerations.

\section{Related Work}
Theoretical background on techniques and algorithms widely adopted in this work relies on several Computer Science and
Applied Mathematics fields such as Algorithms and Data Structures and Artificial Intelligence. In the setting of Web
data extraction, especially algorithms on (DOM) trees play a predominant role. Approaches to analyze similarities
between trees were developed starting from the well-known problem of finding the longest common subsequence(s)
between two strings. Several algorithms were suggested, for example Hirshberg \cite{Hirschberg1975} provided the proof
of correctness of three of them.

Soon, a strong interconnection between this problem and the similarity between trees has been pointed out: Tai
\cite{Tai1979} introduced the notion of \emph{distance} as measure of the (dis)similarity between two trees and
extended the notion of longest common subsequence(s) between strings to trees. Several \emph{tree edit distance}
algorithms were suggested, providing a way to transform a labeled tree in another one through local operations, like
inserting, deleting and relabeling nodes. Bille \cite{Bille2005} reported, in a comprehensive survey on the tree edit
distance and related problems, summarizing approaches and analyzing algorithms.

Algorithms based on the tree edit distance usually are complex to be implemented and imply a high computational cost.
They also provide more information than needed, if one just wants to get an estimate on the similarity. Considering
these reasons, Selkow \cite{StanleyM.Selkow1977} developed a top-down trees isomorphism algorithm called \emph{simple
tree matching}, that establishes the degree of similarity between two trees, analyzing subtrees recursively. Yang
\cite{Yang1991} suggested an improvement of the simple tree matching algorithm, introducing weights.

During years, some improvements to tree edit distance techniques have been introduced: Shasha and Zhang
\cite{Zhang1989} provided proof of correctness and implementation of some new parallelizable algorithms for computing
edit distances between trees, lowering complexity of $\mathsf{O(|T_1| \cdot |T_2|} \cdot$
$\mathsf{min(\emph{depth}(T_1), \emph{leaves}(T_1)) \cdot}$ $\mathsf{min(\emph{depth}(T_2), \emph{leaves}(T_2)))}$, for
the non parallel implementation, to $\mathsf{O(|T_1|+|T_2|)}$, for the parallel one; Klein \cite{Klein1998}, finally,
suggested a fast method for computing the edit distance between unrooted ordered trees in $\mathsf{O(n^3\log n)}$. An
overview of interesting applications of these algorithms in Computer Science can be found in Tekli et al.
\cite{Tekli2009}.

\bigskip

Literature on Web data extraction is manifold: Ferrara et al. \cite{Baumgartner2010} provided a comprehensive survey on
application areas and used techniques, and Laender et al.\ \cite{Laender2002} give a very good overview on wrapper
generation techniques. Focusing on \emph{wrapper adaptation}, Chidlovskii \cite{Chidlovskii2001} presented some
experimental results of combining and applying some grammatical and logic-based rules. Lerman et al. \cite{Lerman2003}
developed a machine-learning based system for wrapper verification and reinduction in case of failure in extracting
data from Web pages.

Meng et al. \cite{20} suggested a new approach, called SG-WRAM (Schema-Guided WRApper Maintenance), for wrapper maintenance, considering that changes in Web pages always preserve syntactic features (i.e. data patterns, string lengths, etc.), hyperlinks and annotations (e.g. descriptive information representing the semantic meaning of a piece of information in its context).

Wong \cite{Wong} developed a probabilistic framework to adapt a previously learned wrapper to unseen Web pages, including the possibility of discovering new attributes, not included in the first one, relying on the extraction knowledge related to the first wrapping task and on the collection of items gathered from the first Web page.

Raposo et al. \cite{Raposo2005} already suggested the possibility of exploiting previously acquired information, e.g. queries results, to re-induct a new wrapper from an old one not working anymore, because of structural changes in Web pages.
Kim et al. \cite{Kim2007} compared results of simple tree matching and a modified weighed version of the same algorithm, in extracting information from HTML Web pages; this approach shares similarities to the one followed here to perform adaptation of wrappers.
Kowalkiewicz et al. \cite{Kowalkiewicz2006} focused on robustness of wrappers exploiting absolute and relative XPath queries.

%%%%%%%%%%%%%%%%%%%%%%%%%%%%
\section{Wrapper Adaptation}

\subsection{Primary Goals}
As previously mentioned, our idea is to compare some helpful structural information stored by applying the wrapper on the original version of the Web page, searching for similarities in the new one.  Regardless of the method of extraction implemented by the wrapping system (e.g. we can consider a simple XPath), elements identified and represented as subtrees of the DOM tree of the Web page, can be exploited to find similarities between two different versions.

\begin{figure*}%
\begin{minipage}{9cm}
    \small (A) /html[1]/body[1]/table[1]/tr[1]/td[1]
\end{minipage}
\begin{minipage}{9cm}
    \small (B) /html[1]/body[1]/table[1]/tr[2]/td
\end{minipage}
\\

\Tree [.{\bf html} [.head ] [.{\bf body} [.{\bf table} [.{\bf tr} [.{\bf td\\$\vdots$} ] ] [.tr [.td ] [.td ] [.td ] ] [.tr [.{td\\$\vdots$} ] [.{td\\$\vdots$} ] ] ]  ] ]
\Tree [.{\bf html} [.head ] [.{\bf body} [.{\bf table} [.tr [.{td\\$\vdots$} ] ] [.{\bf tr} [.{\bf td} ] [.{\bf td} ] [.{\bf td} ] ] [.tr [.{td\\$\vdots$} ] [.{td\\$\vdots$} ] ] ]  ] ]

\caption{Examples of XPaths over trees, selecting one (A) or multiple (B) items}%
\label{ex1}%
\end{figure*}

\bigskip

In the simplest case, the XPath identifies just a single element on the Web page (Figure 1.A); our idea is to look for some elements, in the new Web page, sharing similarities with the original one, evaluating comparable features (e.g. subtrees, attributes, etc.); we call these elements \emph{candidates}; among candidates, the one showing the higher degree of similarity, probably, represents the new version of the original element.

It is possible to extend the same approach in the common case in which the XPath identifies multiple similar elements
on the original page (e.g. a XPath selecting results of a search in a retail online shop, represented as table rows,
divs or list items) (Figure 1.B); it is possible to identify multiple elements sharing a similar structure in the new
page, within a custom level of accuracy (e.g. establishing a threshold value of similarity). Section 4 discusses also
these cases.

Once identified, elements in the new version of the Web page can be extracted as usual, for example just re-inducting
the XPath. Our purpose is to define some rules to enable the wrapper to face the problem of automatically adapting
itself to extract information from the new Web page.

We implemented this approach in a commercial tool \footnote{Lixto Suite, www.lixto.com}; the most efficient way to
acquire some structural information about elements the original wrapper extracts, is to store them inside the
definition of the wrapper itself. For example, generating \emph{signatures} representing the DOM subtree of 
extracted elements from the original Web page, stored as a tree diagram, or a simple XML document (or, even, the HTML itself).
This shrewdness avoids that we need to store the whole original page, ensuring better performances and efficiency.

This technique requires just a few settings during the definition of the wrapper step: the user enables the automatic wrapper adaptation feature and set an accuracy threshold. During the execution of the wrapper, if some XPath definition does not match a node, the wrapper adaptation algorithm automatically starts and tries to find the new version of the missing node.

%%%%%%%%%%%%%%%%%%%%
\subsection{Details}
First of all, to establish a measure of similarity between two trees we need to find some comparable properties between
them. In HTML Web pages, each node of the DOM tree represents an HTML element defined by a tag (or, otherwise, free
text). The simplest way to evaluate similarity between two elements is to compare their \emph{tag name}. Elements own
some particular common attributes (e.g. \emph{id}, \emph{class}, etc.) and some type-related attributes (e.g.
\emph{href} for anchors, \emph{src} for images, etc.); it is possible to exploit this information for additional checks
and comparisons.

The algorithm selects candidates between subtrees sharing the same root element, or, in some cases, \emph{comparable} -but not identical- elements, analyzing tags. This is very effective in those cases of deep modification of the structure of an object (e.g. conversion of tables in divs).

As discussed in Section 2, several approaches have been developed to analyze similarities between HTML trees; for our
purpose we improved a version of the \emph{simple tree matching} algorithm, originally led by Selkow
\cite{StanleyM.Selkow1977}; we call it \emph{clustered tree matching}.

There are two important novel aspects we are introducing in facing the problem of the automatic wrapper adaptation:
first of all, exploiting previously acquired information through a smart and focused usage of the tree similarity
comparison; thus adopting a consolidated approach in a new field of application. Moreover, we contributed applying some
particular and useful changes to the algorithm itself, improving its behavior in the HTML trees similarity measurement.

%%%%%%%%%%%%%%%%%%%%%%%%%%%%%%%%%
\subsection{Simple Tree Matching}
Advantages of adopting this algorithm, which has been shown quite effective for Web data extraction \cite{Kim2007,19},
are multiple; for example, the \emph{simple tree matching} algorithm evaluates similarity between two trees by
producing the maximum matching through dynamic programming, without computing inserting, relabeling and deleting
operations; moreover, tree edit distance algorithms relies on complex implementations to achieve good performance,
instead \emph{simple tree matching}, or similar algorithms are very simple to implement. The computational cost is
$\mathsf{O(n^2 \cdot max(\emph{leaves}(T^{'}), \emph{leaves}(T^{''}))}$ $\mathsf{\cdot max(\emph{depth}(T^{'}),
\emph{depth}(T^{''})))}$, thus ensuring good performances, applied to HTML trees. There are some limitations; most of
them are irrelevant but there is an important one: this approach can not match permutations of nodes. Despite this
intrinsic limit, this technique appears to fit very well to our purpose of measuring HTML trees similarity.

Let \emph{d(n)} to be the degree of a node \emph{n} (i.e. the number of first-level children); let T(i) to be the i-\emph{th} subtree of the tree rooted at node T; this is a possible implementation of the \emph{simple tree matching} algorithm:

\begin{algorithm}[H]
\caption{SimpleTreeMatching($T^{'}$, $T^{''}$)}
\label{alg1}
\begin{algorithmic}[1]
    \IF{$T^{'}$ has the same label of $T^{''}$}
        \STATE $m \leftarrow$ $d(T^{'})$
        \STATE $n \leftarrow$ $d(T^{''})$
        \FOR{$i = 0$ to $m$}
            \STATE $M[i][0] \leftarrow 0$;
        \ENDFOR
        \FOR{$j = 0$ to $n$}
            \STATE $M[0][j] \leftarrow 0$;
        \ENDFOR
        \FORALL{$i$ such that $1\leq i\leq m$}
            \FORALL{$j$ such that $1\leq j \leq n$}
                \STATE $M[i][j] \leftarrow$ Max($M[i][j-1]$, $M[i-1][j]$, $M[i-1][j-1] + W[i][j]$) where $W[i][j]$ = SimpleTreeMatching($T^{'}(i-1)$, $T^{''}(j-1)$)
            \ENDFOR
        \ENDFOR
        \STATE return M[m][n]+1
    \ELSE
        \STATE return 0
    \ENDIF
\end{algorithmic}
\end{algorithm}

%%%%%%%%%%%%%%%%%%%%%%%%%%%%%%%%%%%%
\subsection{Clustered Tree Matching}
 Let $t(n)$ to be the number of total siblings of a node \emph{n} including itself:
\begin{algorithm}[H]
\caption{ClusteredTreeMatching($T^{'}$, $T^{''}$)}
\label{alg2}
\begin{algorithmic}[1]
    \STATE \COMMENT{Change line 11 with the following code}
    \IF{$m > 0$ AND $n > 0$}
        \STATE return M[m][n] * 1 / Max($t(T^{'})$, $t(T^{''})$)
    \ELSE
        \STATE return M[m][n] + 1 / Max($t(T^{'})$, $t(T^{''})$)
    \ENDIF
\end{algorithmic}
\end{algorithm}

In order to better reflect a good measure of similarity between HTML trees, we applied some focused changes to the way of assignment of a value for each matching node.

In the \emph{simple tree matching} algorithm the assigned matching value is always 1. After leading some analysis and considerations on structure of HTML pages, our intuition was to assign a weighed value, with the purpose of attributing less importance to slight changes, in the structure of the tree, when they occur in deep sublevels (e.g. missing/added leaves, small truncated/added branches, etc.) and also when they occur in sublevels with many nodes, because these mainly represent HTML list of items, table rows, etc., usually more likely to modifications.

In the \emph{clustered tree matching}, the weighed value assigned to a match between two nodes is 1, divided by the greater number of siblings with respect to the two compared nodes, considering nodes themselves (e.g. Figure 2.A, 2.B); thus reducing the impact of missing/added nodes.

Before assigning a weight, the algorithm checks if it is comparing two leaves or a leaf with a node which has children (or two nodes which both have children). 
The final contribution of a sublevel of leaves is the sum of assigned weighted values to each leaf (cfr. Code Line (4,5)); thus, the contribution of the parent node of those leaves is equal to its weighed value multiplied by the sum of contributions of its children (cfr. Code Line (2,3)). 
This choice produces an effect of \emph{clustering} the process of matching, subtree by subtree; this implies that, for each
sublevel of leaves the maximum sum of assigned values is 1; thus, for each parent node of that sublevel the maximum
value of the multiplication of its contribution with the sum of contributions of its children, is 1; each cluster,
singly considered, contributes with a maximum value of 1. In the last recursion of this top-down algorithm, the two
roots will be evaluated. The resulting value at the end of the process is the measure of similarity between the two
trees, expressed in the interval [0,1]. The closer the final value is to 1, the more the two trees are similar.

\begin{figure*}
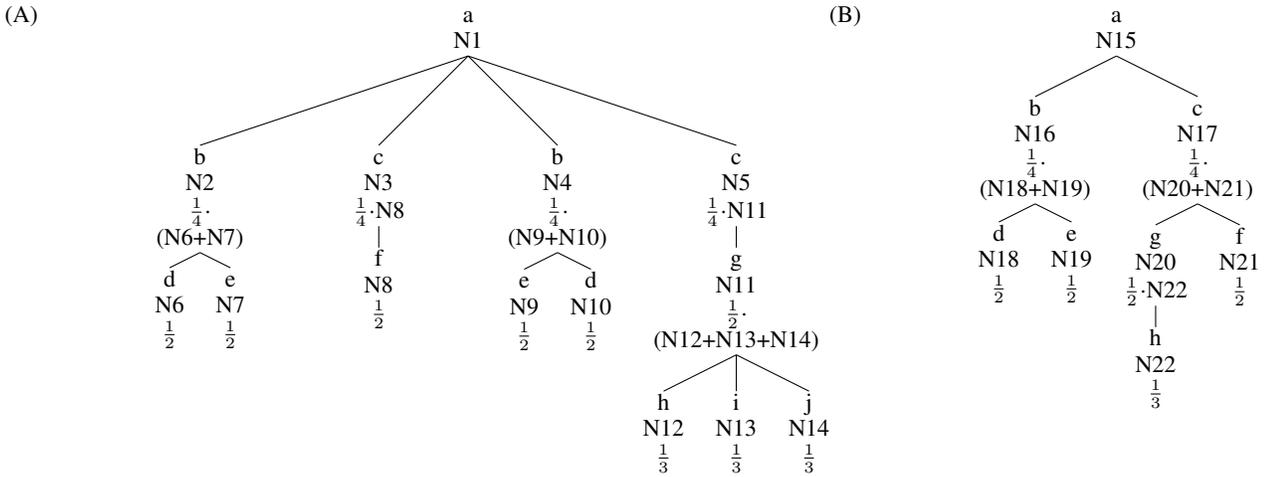


\small (A)
\Tree [.{a\\\small N1} [.{b\\\small N2\\$\frac{1}{4}\cdot$\\\small (N6+N7)} [.{d\\\small N6\\ $\frac{1}{2}$} ] [.{e\\\small N7\\ $\frac{1}{2}$} ] ] [.{c\\\small N3\\$\frac{1}{4}\cdot$\small N8} [.{f\\\small N8\\ $\frac{1}{2}$} ]  ] [.{b\\\small N4\\$\frac{1}{4}\cdot$\\\small (N9+N10)} [.{e\\\small N9\\ $\frac{1}{2}$} ] [.{d\\\small N10\\ $\frac{1}{2}$} ] ] [.{c\\\small N5\\$\frac{1}{4}\cdot$\small N11} [.{g\\\small N11\\$\frac{1}{2}\cdot$\\\small (N12+N13+N14)} [.{h\\\small N12\\ $\frac{1}{3}$} ] [.{i\\\small N13\\ $\frac{1}{3}$} ] [.{j\\\small N14\\ $\frac{1}{3}$} ] ] ] ]
\small (B)
\Tree [.{a\\\small N15} [.{b\\\small N16\\$\frac{1}{4}\cdot$\\\small (N18+N19)} [.{d\\\small N18\\ $\frac{1}{2}$} ] [.{e\\\small N19\\ $\frac{1}{2}$} ]  ] [.{c\\\small N17\\$\frac{1}{4}\cdot$\\\small (N20+N21)} [.{g\\\small N20\\$\frac{1}{2}\cdot$\small N22} [.{h\\\small N22\\ $\frac{1}{3}$} ] ] [.{f\\\small N21\\ $\frac{1}{2}$} ] ] ]

\normalsize
\bigskip

\label{ex2}
\caption{\emph{A} and \emph{B} are two similar labeled rooted trees}

\end{figure*}

Let us analyze the behavior of the algorithm with an example, already used by \cite{Yang1991} and \cite{19} to explain
the simple tree matching (Figure 2): 2.A and 2.B are two very simple generic rooted labeled trees (i.e. the same
structure of HTML trees). They show several similarities except for some missing nodes/branches. Applying the
\emph{clustered tree matching} algorithm, in the first step (Figure 2.A, 2.B) contributions assigned to leaves, with
respect to matches between the two trees, reflect the past considerations (e.g. a value of $\frac{1}{3}$ is established
for nodes (h), (i) and (j), although two of them are missing in 2.B). Going up to parents, the summation of
contributions of matching leaves is multiplied by the relative value of each node (e.g. in the first sublevel, the
contribution of each node is $\frac{1}{4}$ because of the four first-sublevel nodes in 2.A).

Once completed these operations for all nodes of the sublevel, values are added and the final measure of similarity for the two trees is obtained. Intuitively, in more complex and deeper trees, this process is iteratively executed for all the sublevels. The deeper a mismatch is found, the less its missing contribution will affect the final measure of similarity. Analogous considerations hold for missing/added nodes and branches, sublevels with many nodes, etc. Table 1 shows M and W matrices containing contributions and weights.

In this example, ClusteredTreeMatching(2.A, 2.B) returns a measure of similarity of $\frac{3}{8}$ ($0.375$) whereas SimpleTreeMatching(2.A, 2.B) would return a mapping value of 7; the main difference on results provided by these two algorithms is the following: our \emph{clustered tree matching} intrinsically produces an absolute measure of similarity between the two compared trees; the \emph{simple tree matching}, instead, returns the mapping value and then it needs subsequent operations to establish the measure of similarity.

Hypothetically, in the \emph{simple tree matching} case, we could suppose to establish a good estimation of similarity dividing the mapping value by the total number of nodes of the tree with more nodes; indeed, a value calculated in this way would be linear with respect to the number of nodes, thus ignoring important information as the position of mismatches, the number of mismatches with respect to the total number of subnodes/leaves in a particular sublevel, etc.

In this case, for example, the measure of similarity between 2.A and 2.B, applying this approach, would be $\frac{7}{14}$ (0.5). A greater value of similarity could suggest, wrongly, that this approach is more accurate. Experimentation showed us that, the closer the measure of similarity is to reflect changes in complex structures, the higher the accuracy of the matching process is. This fits particularly well for HTML trees, which often show very rich and articulated structures.

The main advantage of using the \emph{clustered tree matching} algorithm is that, the more the structure of considered trees is complex and similar, the more the measure of similarity will be accurate. On the other hand, for simple and quite different trees the accuracy of this approach is lower than the one ensured by the \emph{simple tree matching}. But, as already underlined, the most of changes in Web pages are usually minor changes, thus \emph{clustered tree matching} appears to be a valid technique to achieve a reliable process of automatic wrapper adaptation.

\begin{table}[!h]
\begin{minipage}{4cm}
\begin{tabular}{|@{} c @{}| c | c |}
     \hline
     W  & \small N18 & \small N19\\
     \hline
     \small N6  & \cellcolor[gray]{0.8}$\frac{1}{2}$  & 0\\
     \hline
     \small N7  & 0 & \cellcolor[gray]{0.8}$\frac{1}{2}$\\
     \hline
    \end{tabular}
\end{minipage}
\begin{minipage}{4cm}
\begin{tabular}{| c | c | c | c |}
     \hline
     M  & 0 & \small N18 & \small N18-19\\
     \hline
     0  & 0 & 0 & 0\\
     \hline
     \small N6  & 0 & \cellcolor[gray]{0.8}$\frac{1}{2}$  & $\frac{1}{2}$\\
     \hline
     \small N6-7    & 0 & $\frac{1}{2}$  & \cellcolor[gray]{0.8}1\\
     \hline
    \end{tabular}
\end{minipage}
\end{table}

\begin{table}[!h]
\begin{minipage}{4cm}
\begin{tabular}{|@{} c @{}| c | c |}
     \hline
     W  & \small N18 & \small N19\\
     \hline
     \small N9  & 0 & \cellcolor[gray]{0.8}$\frac{1}{2}$\\
     \hline
     \small N10 & \cellcolor[gray]{0.8}$\frac{1}{2}$  & 0\\
     \hline
    \end{tabular}
\end{minipage}
\begin{minipage}{4cm}
    \begin{tabular}{| c | c | c | c |}
     \hline
     M  & 0 & \small N18 & \small N18-19\\
     \hline
     0  & 0 & 0 & 0\\
     \hline
     \small N9  & 0 & 0 & \cellcolor[gray]{0.8}$\frac{1}{2}$\\
     \hline
     \small N9-10   & 0 & \cellcolor[gray]{0.8}$\frac{1}{2}$  & \cellcolor[gray]{0.8}$\frac{1}{2}$\\
     \hline
    \end{tabular}
\end{minipage}
\end{table}

\begin{table}[!h]
\begin{minipage}{4cm}
    \begin{tabular}{|@{} c @{}| c |}
        \hline
         W  & \small N8\\
         \hline
         \small N20 & 0\\
         \hline
         \small N21 & \cellcolor[gray]{0.8}$\frac{1}{2}$\\
         \hline
        \end{tabular}
\end{minipage}
\begin{minipage}{4cm}
    \begin{tabular}{| c | c | c |}
        \hline
         M  & 0 & \small N8\\
         \hline
         0  & 0 & 0\\
         \hline
         \small N20 & 0 & 0\\
         \hline
         \small N20-21  & 0 & \cellcolor[gray]{0.8}$\frac{1}{2}$\\
         \hline
        \end{tabular}
\end{minipage}
\end{table}

\begin{table}[!h]
\begin{minipage}{4cm}
    \begin{tabular}{|@{} c @{}| c | c | c |}
            \hline
             W  & \small N12 & \small N13 & \small N14\\
             \hline
             \small N22 & \cellcolor[gray]{0.8}$\frac{1}{3}$ & 0 & 0\\
             \hline
            \end{tabular}
\end{minipage}
\begin{minipage}{4cm}
    \begin{tabular}{| c | c | c | @{}c@{} | @{}c@{} |}
    \hline
     M  & 0 & \small N12 & \small N12-13 & \small N12-14\\
     \hline
     0  & 0 & 0 & 0 & 0\\
     \hline
     \small N22 & 0 & \cellcolor[gray]{0.8}$\frac{1}{3}$ & $\frac{1}{3}$  & $\frac{1}{3}$\\
     \hline
    \end{tabular}
\end{minipage}
\end{table}

\begin{table}[!h]
\begin{minipage}{4cm}
    \begin{tabular}{|@{} c @{}| c |}
        \hline
         W  & \small N11\\
         \hline
         \small N20 & \cellcolor[gray]{0.8}$\frac{1}{6}$\\
         \hline
         \small N21 & 0\\
         \hline
        \end{tabular}
\end{minipage}
\begin{minipage}{4cm}
    \begin{tabular}{| c | c | c |}
        \hline
         M  & 0 & \small N11\\
         \hline
         0  & 0 & 0\\
         \hline
         \small N20 & 0 & \cellcolor[gray]{0.8}$\frac{1}{6}$\\
         \hline
         \small N20-21  & 0 & \cellcolor[gray]{0.8}$\frac{1}{6}$\\
         \hline
        \end{tabular}
\end{minipage}
\end{table}

\begin{table}[!h]
\begin{minipage}{4cm}
    \begin{tabular}{|@{} c @{}| @{}c@{} | c | c | c | c |}
        \hline
         W  & \small N2 & \small N3 & \small N4 & \small N5\\
         \hline
         \small N16 & $\frac{1}{4}$ & 0 &  \cellcolor[gray]{0.8}$\frac{1}{8}$  & 0\\
         \hline
         \small N17 & 0 & \cellcolor[gray]{0.8}$\frac{1}{8}$ & 0 & \cellcolor[gray]{0.8}$\frac{1}{24}$ \\
         \hline
        \end{tabular}
\end{minipage}
\begin{minipage}{4cm}
    \begin{tabular}{|@{}c@{}|c|@{}c@{}|@{}c@{}|@{}c@{}|c|}
        \hline
         M & 0 & \small N2 & \small N2-3 & \small N2-4 & \small N2-5\\
         \hline
         0  & 0 & 0 & 0 & 0 & 0\\
         \hline
         \small N16 & 0 & $\frac{1}{4}$ & $\frac{1}{4}$  &  $\frac{1}{4}$ & $\frac{1}{4}$ \\
         \hline
         \small N16-17  & 0 & $\frac{1}{4}$  & $\frac{3}{8}$ & $\frac{3}{8}$ & \cellcolor[gray]{0.8}$\frac{3}{8}$ \\
         \hline
        \end{tabular}
\end{minipage}
\label{tab1}
\caption{\emph{W} and \emph{M} matrices for matching subtrees}
\end{table}

\normalsize
%%%%%%%%%%%%%%%%%%%%%%%%%
\section{Experimentation}
In this section we discuss some experimentation performed on common fields of application \cite{Baumgartner2010} and following results. We tried to automatically adapt wrappers, previously built to extract information from particular Web pages, after some -often minor- structural changes. All the followings are real use cases: we did not modify any Web page, original owners did; thus re-publishing pages with changes and altering the behavior of old wrappers. Our will to handle real use cases limits the number of examples of this study. These real use cases confirmed our expectations and simulations on ad hoc examples we prepared to test the algorithms.

We obtained an acceptable degree of precision using the \emph{simple tree matching} and a great rate of precision/recall using the \emph{clustered tree matching}. Precision, Recall and F-Measure will summarize these results showed in Table 2. We focused on following areas, widely interested by Web data extraction:

\begin{itemize}
    \item News and Information: Google News \footnote{http://news.google.com} is a valid use case for wrapper adaptation; templates change frequently and sometimes is not possible to identify elements with old wrappers.

    \item Web Search: Google Search \footnote{http://www.google.com} completely rebuilt the results page layout in the same period we started our experimentation \footnote{http://googleblog.blogspot.com/2010/05/spring-metamorphosis-googles-new-look.html}; we exploited the possibility of automatically adapting wrappers built on the old version of the \emph{result page}.

    \item Social Networks: another great example of continuous restyling is represented by the most common social network, Facebook \footnote{http://www.facebook.com}; we successfully adapted wrappers extracting friend lists also exploiting additional checks performed on attributes.

    \item Social Bookmarking: building \emph{folksonomies} and \emph{tagging} contents is a common behavior of Web 2.0 users. Several Websites provide platforms to aggregate and classify sources of information and these could be extracted, so, as usual, wrapper adaptation is needed to face chages. We choose Delicious \footnote{http://www.delicious.com} for our experimentation obtaining stunning results.

    \item Retail: these Websites are common fields of application of data extraction and Ebay \footnote{http://www.ebay.com} is a nice real use case for wrapper adaptation, continuously showing, often almost invisible, structural changes which require wrappers to be adapted to continue working correctly.

    \item Comparison Shopping: related to the previous category, many Websites provide tools to compare prices and features of products. Often, it is interesting to extract this information and sometimes this task requires adaptation of wrappers to structural changes of Web pages. Kelkoo \footnote{http://shopping.kelkoo.co.uk} provided us a good use case to test our approach.

    \item Journals and Communities: Web data extraction tasks can also be performed on the millions of online Web journals, blogs and forums, based on open source blog publishing applications (e.g. Wordpress \footnote{http://wordpress.org}, Serendipity \footnote{http://www.s9y.org}, etc.), CMS (e.g. Joomla \footnote{http://www.joomla.org}, Drupal \footnote{http://drupal.org}, etc.) and community management systems (e.g. phpBB \footnote{http://www.phpbb.com}, SMF \footnote{http://www.simplemachines.org}, etc.). These platforms allow changing templates and often this implies wrappers must be adapted. We lead the automatic adaptation process on Techcrunch \footnote{http://www.techcrunch.com}, a tech journal built on Wordpress.

\end{itemize}

We adapted wrappers for these 7 use cases considering 70 Web pages; Table 2 summarizes results obtained comparing the two algorithms applied on the same page, with the same configuration (threshold, additional checks, etc.). \emph{Threshold} (thr. in Table 2) represents the value of similarity required to match two trees. The columns \emph{tp}, \emph{fp} and \emph{fn} represent true and false positive and false negative items extracted from Web pages through adapted wrappers.

In some situations of deep changes (Facebook, Kelkoo, Delicious) we had to lower the threshold in order to correctly
match the most of the results. Both the algorithms show a great elasticity and it is possible to adapt wrappers with a
high degree of reliability; the \emph{simple tree matching} approach shows a weaker recall value, whereas performances
of the \emph{clustered tree matching} are stunning (F-Measure greater than 98\% is an impressive result). Sometimes,
additional checks on nodes attributes are performed to refine results of both the two algorithms. For example, we can
additionally include attributes as part of the node label (e.g. \emph{id}, \emph{name} and \emph{class}) to refine
results. Also without including these additional checks, the most of the time the false positive results are very
limited in number (cfr. the Facebook use case).

Figure 3 shows a screenshot of the developed tool, performing an automatic wrapper adaptation task: in this example we adapted the wrapper defined for extracting Google news, whereas the original XPath was not working because of some structural changes in the layout of news. 
Elements identified by the original XPath are highlighted in red in the upper browser, elements highlighted in the bottom browser represent the recognized ones through the wrapper adaptation process. 

\begin{table}
        \begin{tabular}{| c | c | c | c | c | c | c | c |}
    \cline{3-8}
        \multicolumn{2}{r|}{} & \multicolumn{3}{c|}{\small Simple T. M.} & \multicolumn{3}{c|}{\small Clustered T. M.} \\
    \cline{3-8}
        \multicolumn{2}{r|}{} & \multicolumn{3}{c|}{Prec./Rec.} & \multicolumn{3}{c|}{Prec./Rec.}\\
    \hline
        URL & thr. & tp & fp & fn & tp & fp & fn\\
    \hline
        news.google.com & 90\% & 604 & - & 52 &  644 & - & 12\\
    \hline
        google.com & 80\% & 100 & - & 60 &  136 & - & 24 \\
    \hline
        facebook.com & 65\% & 240& 72 & - &  240&12 & - \\
    \hline
        delicious.com & 40\% & 100 & 4 & - & 100 & - & - \\
    \hline
        ebay.com & 85\% & 200 & 12 & - &  196 & -  & 4 \\
    \hline
        kelkoo.co.uk & 40\% & 60 & 4 & - & 58 & - & 2 \\
    \hline
        techcrunch.com & 85\% &  52 & - & 28 &  80 & - & - \\
    \hline
        \textbf{Total} & \cellcolor[gray]{0.8} & 1356 & 92 & 140 & 1454 & 12 & 42\\
    \hline
        \textbf{Recall} & \cellcolor[gray]{0.8} & \multicolumn{3}{r|}{90.64\%} & \multicolumn{3}{r|}{97.19\%}\\
    \hline
        \textbf{Precision} & \cellcolor[gray]{0.8} & \multicolumn{3}{r|}{93.65\%} & \multicolumn{3}{r|}{99.18\%}\\
    \hline
        \textbf{F-Measure} & \cellcolor[gray]{0.8} & \multicolumn{3}{r|}{92.13\%} & \multicolumn{3}{r|}{98.18\%}\\
		\hline
	\end{tabular}
\label{tab2}
\caption{Experimental results}
\end{table}

%%%%%%%%%%%%%%%%%%%%%%%

\begin{figure*}%
	\includegraphics[width=520pt]{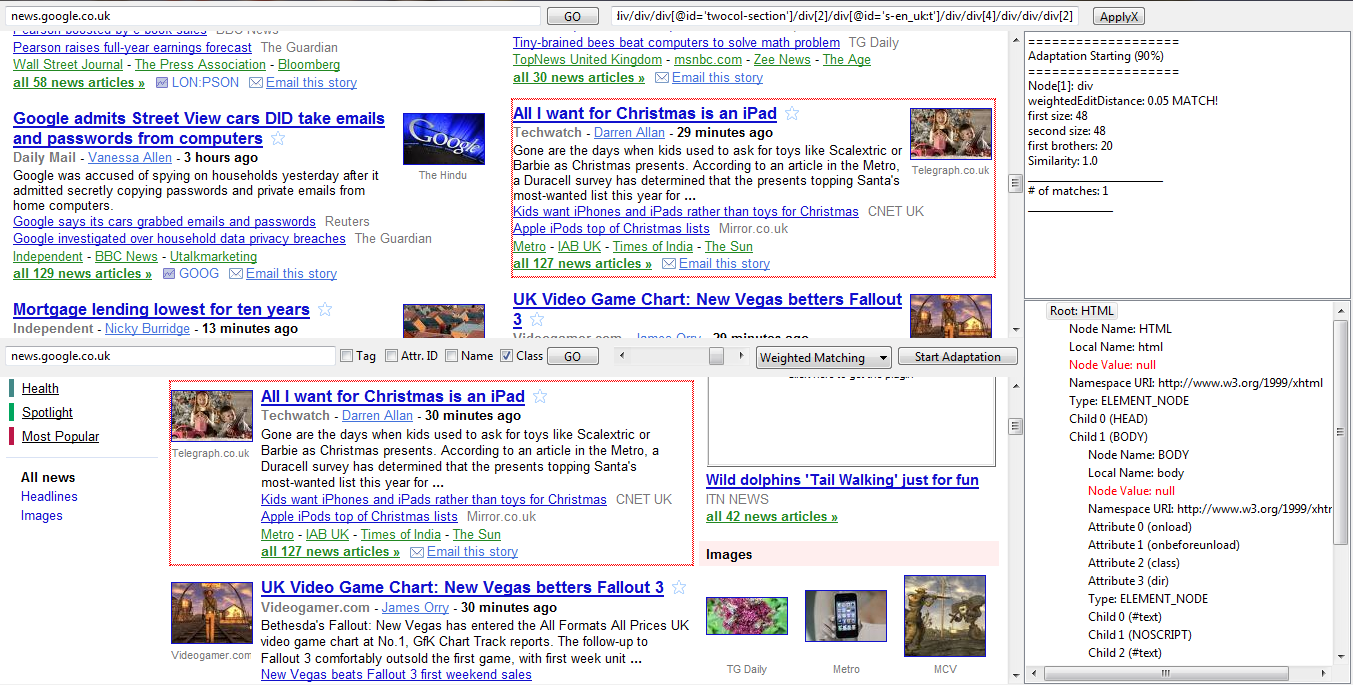}%
	\caption{An example of Wrapper Adaptation}%
	\label{fig3}%
\end{figure*}

%%%%%%%%%%%%%%%%%%%%%%%%%%%%%%%%%%%%
\section{Conclusion}
This work presents new scenarios, analyzing Wrapper Adaptation related problems from a novel point of view, introducing improvements to algorithms and new fields of application.

There are several possible improvements to our approach we can already imagine. First of all, it could be very interesting to extend the matching criteria we used, making the tree matching algorithm smarter. Actually, we already included features like analyzing attributes (e.g. \emph{id}, \emph{name} and \emph{class}) instead of just comparing labels/tags or node types. The accuracy of the matching process benefits of these additional checks and it is possible, for example, to improve this aspect with a more complex matching technique, containing full path information, all attributes, etc.

It could be interesting to compare these algorithms, with other tree edit distance approaches working with
permutations; although, intuitively, \emph{simple tree matching} based algorithms can not handle permutations on nodes,
maybe it is possible to develop some enhanced version which solves this limitation. Furthermore, just considering the
tree structure can be limiting in some particular situations: if a new node has only empty textual fields (or, equally,
if a deleted node had only empty fields) we could suppose its weight should be null. In some particular situation this
inference works well, in some others, instead, it could provoke mismatches. It could also be interesting to exploit
textual properties, nevertheless, not necessarily adopting Natural Language Processing techniques (e.g. using
logic-based approaches, like regular expressions, or string edit distance algorithms, or just the length of strings --
treating two nodes as equal only if the textual content is similar or of similar length).

The tree grammar could also be used in a machine learning approach, for example creating some tree templates to match similar structures or tree/cluster diagrams to classify and identify several different topologies of common substructures in Web pages. This process of simplification is already used to store a light-weight snapshot of elements identified by a wrapper applied on a Web page, at the time of extraction; actually, this feature allows the algorithm to work also without the original version of the page, but just exploiting some information about extracted items. This possibility opens new scenarios for future work on Wrapper Adaptation.

Concluding, the \emph{clustered tree matching} algorithm we described is very extensible and resilient, so as ensuring its use in several different fields, for example it perfectly fits in identifying similar elements belonging to a same structure but showing some small differences among them. Experimentation on wrapper adaptation has already been performed inside a productive tool, the Lixto Suite, this because our approach has been shown to be solid enough to be implemented in real systems, ensuring great reliability and, generically, stunning results.

% conference papers do not normally have an appendix

% use section* for acknowledgement
%\section*{Acknowledgment}

%The authors would like to thank...

% trigger a \newpage just before the given reference
% number - used to balance the columns on the last page
% adjust value as needed - may need to be readjusted if
% the document is modified later
%\IEEEtriggeratref{8}
% The "triggered" command can be changed if desired:
%\IEEEtriggercmd{\enlargethispage{-5in}}

% references section

% can use a bibliography generated by BibTeX as a .bbl file
% BibTeX documentation can be easily obtained at:
% http://www.ctan.org/tex-archive/biblio/bibtex/contrib/doc/
% The IEEEtran BibTeX style support page is at:
% http://www.michaelshell.org/tex/ieeetran/bibtex/
%\bibliographystyle{IEEEtran}
% argument is your BibTeX string definitions and bibliography database(s)
%\bibliography{IEEEabrv,../bib/paper}
%
% <OR> manually copy in the resultant .bbl file
% set second argument of \begin to the number of references
% (used to reserve space for the reference number labels box)
%\begin{thebibliography}{1}

%\bibitem{IEEEhowto:kopka}
%H.~Kopka and P.~W. Daly, \emph{A Guide to \LaTeX}, 3rd~ed.\hskip 1em plus  0.5em minus 0.4em\relax Harlow, England: Addison-Wesley, 1999.

%\end{thebibliography}

%%%%%%%%%%%%%%%%%%%%%%%%%%%%%%
\bibliographystyle{IEEEtran}
\bibliography{IEEEabrv,biblio}

% that's all folks
\end{document}